\documentclass{article}
\usepackage{spconf,amsmath,graphicx}

\usepackage{tikz}
\usepackage[utf8]{inputenc}
\usepackage[T1]{fontenc}
\usepackage{url}
\usepackage{booktabs}
\usepackage{amsfonts}
\usepackage{nicefrac}
\usepackage{microtype}
\usepackage{wrapfig}

\usepackage{amssymb}
\usepackage{bm}
\usepackage{url}
\usepackage{stmaryrd}
\usepackage{mathtools}
\usepackage{calc}
\usepackage{booktabs}
\usepackage{nccmath}

\usepackage{algorithm}
\usepackage{algorithmic}

\usepackage{multirow}
\usepackage{mathtools}
\usepackage{enumerate}
\usepackage{amsthm}
\usepackage[subrefformat=parens]{subcaption}
\usepackage{cite}
\newcommand{\figref}[1]{Fig.~\ref{#1}}
\newcommand{\tabref}[1]{Table~\ref{#1}}
\newcommand{\secref}[1]{Sec.~\ref{#1}}
\newcommand{\ph}{\phantom{$0$}}
\newcommand{\Ph}{\phantom{(0)}}
\newcommand{\PH}{\phantom{0(0)}}
\newcommand{\myeqref}[1]{Eq.~\eqref{#1}}
\usepackage{pifont}
\newcommand{\xmark}{\ding{55}}%
\newcommand{\cmark}{\ding{51}}%

\title{gSwin: Gated MLP Vision Model\\with Hierarchical Structure of Shifted Window}

\name{Mocho Go, Hideyuki Tachibana}
\address{PKSHA Technology, Inc.,
    Hongo, Bunkyo City, Tokyo, Japan}

\usepackage{tikz}
\newcommand\copyrighttext{%
    \footnotesize \copyright 2023 IEEE. Personal use of this material is permitted.
    Permission from IEEE must be obtained for all other uses, in any current or future
    media, including reprinting/republishing this material for advertising or promotional
    purposes, creating new collective works, for resale or redistribution to servers or
    lists, or reuse of any copyrighted component of this work in other works.
}\newcommand\mycopyrightnotice{%
\begin{tikzpicture}[remember picture,overlay]
\node[anchor=north,yshift=-10pt] at 
    (current page.north) {{\parbox{\dimexpr\textwidth-\fboxsep-\fboxrule\relax}{\copyrighttext}}};
\end{tikzpicture}%
}
\begin{document}
\ninept
\maketitle

\begin{abstract}
Following the success in language domain, the self-attention mechanism (Transformer)
has been adopted in the vision domain and achieving great success recently.
Additionally, the use of multi-layer perceptron (MLP) is also explored in the vision domain as another stream.
These architectures have been attracting attention recently to alternate the traditional CNNs,
and many Vision Transformers and Vision MLPs have been proposed.
By fusing the above two streams, this paper proposes gSwin,
a novel vision model which can consider spatial hierarchy and
locality due to its network structure similar to the Swin Transformer,
and is parameter efficient due to its gated MLP-based architecture.
It is experimentally confirmed that the gSwin can achieve better accuracy
than Swin Transformer on three common tasks of vision
with smaller model size.
\end{abstract}

\mycopyrightnotice

\section{Introduction}
Since the great success of AlexNet~\cite{krizhevsky2012imagenet} in ILSVRC2012 ushered in the era of deep learning,
convolutional neural networks (CNNs)
have been dominant in the image domain for the past decade.
On the other hand, looking outside the image domain,
the Transformer architecture based on the stacked multi-head self-attention (MSA)~\cite{vaswani2017attention}
that directly connects one token to another has been proposed in the language domain,
and Transformer-based models such as BERT~\cite{devlin2018bert}
have had great success.
Following the success in the language domain,
there has been active research in recent years on the application of Transformers in the vision domain,
viz.\ the Vision Transformer (ViT)~\cite{dosovitskiy2020image}, with remarkably promising results.

A shortcoming of Transformer is that its multi-head attention is so huge that the module is computationally expensive and data inefficient.
For example, ViT has too large data capacity, and is not adequately trained on a medium-sized dataset such as ImageNet-1K~\cite{deng2009imagenet}.
To remedy such problems, a number of improved ViTs and learning strategies have been proposed.
In this vein, some variants of Transformers have been proposed
which explicitly incorporate the visual hierarchy,
including the Swin Transformer~\cite{liu2021swin},
with the motivation that the scale hierarchy,
an important property of vision, is naturally incorporated in CNNs but not in Transformers (\secref{label:006}).
Another important stream of research has been the model simplification,
in particular a series of methods using only MLPs~\cite{tolstikhin2021mlp,touvron2021resmlp,liu2021pay}.
These methods were proposed based on the question of whether self-attention modules are really necessary
for the vision tasks (\secref{label:007}).

In this paper, we propose \textit{gSwin} which reintegrates the above two pathways that are evolving
in different directions after breaking off from ViT. Our gSwin
is based on the basic framework of gMLP~\cite{liu2021pay} as a parameter-efficient Transformer alternative,
and the Swin Transformer~\cite{liu2021swin} to capture the visual hierarchical structure.
Although it is a simple idea of merging two independent streams, we have experimentally confirmed that it gives promising results.
Our experiments showed that gSwin-T outperformed Swin-T
with
$+0.4$~Top-1 accuracy on ImageNet-1K~\cite{deng2009imagenet},
$+0.5$~box-AP and $+0.4$~mask-AP on COCO~\cite{lin2014microsoft},
and $+1.9$~mIoU on ADE20K~\cite{zhou2019semantic}.
The gSwin-S was also competitive with Swin-S.
Both gSwin models are smaller than their Swin Transformer counterparts.

\section{Related Work}\label{label:000}
\begin{figure*}[t]
	\def\figheight{3.cm}
	\centering
    \subfloat[ViT~\cite{dosovitskiy2020image}]{
        \includegraphics[height=\figheight]{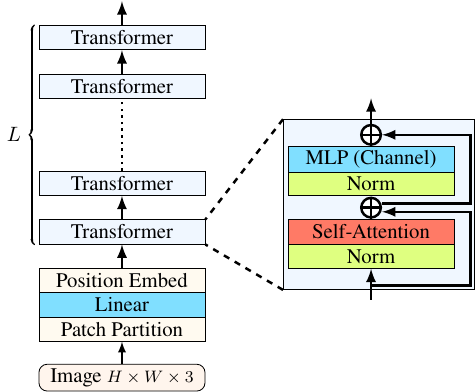}
        \label{label:001}
    }
    \centering
	\,
    \subfloat[gMLP~\cite{liu2021pay}]{
        \includegraphics[height=\figheight]{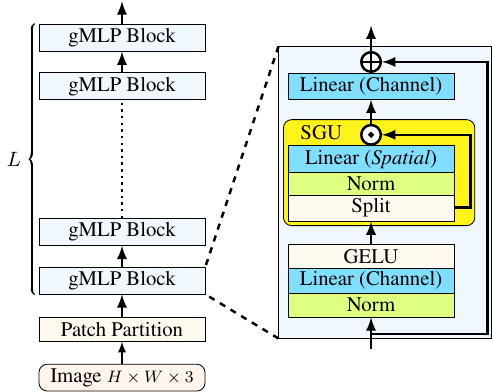}
        \label{label:002}
    }
    \centering
	\,
    \subfloat[Swin Transformer~\cite{liu2021swin}]{
        \includegraphics[height=\figheight]{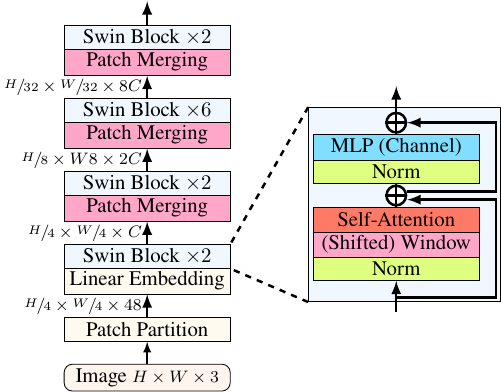}
        \label{label:003}
    }
	\,
    \subfloat[\textbf{gSwin (proposed)}]{
		\includegraphics[height=\figheight]{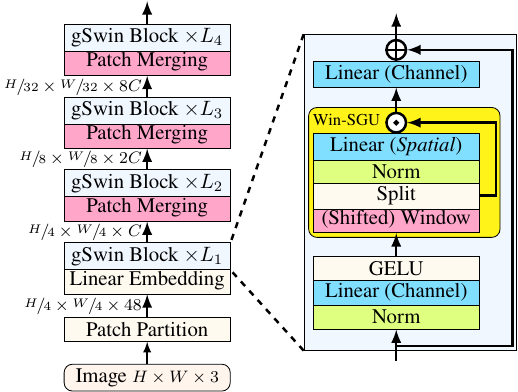}
		\label{label:004}
	}
	\caption{
        Comparison of existing vision models and the proposed method.
    }
    \label{label:005}
\end{figure*}

\subsection{Transformers for Vision}\label{label:006}
Recently, vision models principally built on self-attention mechanisms have been proposed~\cite{carion2020end,wu2020visual,dosovitskiy2020image}.
Of these, ViT~\cite{dosovitskiy2020image} (\figref{label:001}) produced competitive results and marked the beginning of research on vision models based on the Transformer architecture.
In the 2D vision domain, a vast amount of research to improve ViT has been actively conducted.
One trend is oriented toward improving learning strategies,
such as~\cite{touvron2021training,chen2021empirical},
and another direction is the improvement of model architectures.

A major stream of research on improving model structure is the incorporation of spatial structure in images,
which has not been explicitly used in Transformers,
except for the customary use of positional encoding and the patch partitioning before the initial layer.
In specific, auxialiry use of convolution layers~\cite{srinivas2021bottleneck,d2021convit,graham2021levit,wu2020visual,li2021localvit}
and the hierarchical structures~\cite{yuan2021tokens,pan2021scalable,han2021transformer,fan2021multiscale,wang2021pyramid,zhang2021multi}
have been explored.
The latter includes the Swin Transformer~\cite{liu2021swin} (\figref{label:003}).
Particular essentials in the structure of Swin Transformer are the bottom-up hierarchical structure for each resolution and the shifting of windows.
Indeed, it has been reported that as long as this structure is maintained,
the attention can actually be replaced by average pooling (PoolFormer),
or even an identity map, although performance is slightly reduced~\cite{yu2021metaformer}.
This fact would be a evidence that the hierarchical structure of the Swin Transformer is advantageous as a vision model.

\subsection{MLP Models for Vision}\label{label:007}
Aside from Vision Transformers, light weight alternatives of self-attention mechanisms have been explored especially in the NLP domain.
In particular, axis-wise multi-layer perceptron (MLP) architectures have also been attracting attention recently,
including
MLP-Mixer~\cite{tolstikhin2021mlp}, ResMLP~\cite{touvron2021resmlp} and gMLP~\cite{liu2021pay},
which were proposed almost simultaneously.
Following these studies,
a number of vision models based on axis-wise MLPs have been proposed recently~\cite{guo2021hire,hou2022vision,chen2021cyclemlp,yu2022s2,lian2021mlp}.
A method that deserves special mention in a different line is the gMLP~\cite{liu2021pay} (\figref{label:002}).
Its noteworthy characteristic is the use of the spatial gating unit (SGU),
a gating mechanism for spatial mixing, which mimics an attention mechanism.
Another feature of this method is its simplicity: only one linear (fully-connected projection)
layer is used where the stacking of two or more dense layers are required in other related methods.
The gMLP model has achieved competitive performance both in language and vision domains with standard
Transformer models BERT~\cite{devlin2018bert} and ViT~\cite{dosovitskiy2020image}.
It is also reported that the gMLP could learn shift-invariant features.

\section{Methodology}\label{label:008}
\subsection{Overall Network Architecture of gSwin}
In this paper, we aim to achieve high performance while the computational and parameter efficiency is on par with the existing methods, by incorporating both merits of the two methods that have evolved in different directions from ViT~\cite{dosovitskiy2020image} (\figref{label:001}). In particular, we consider the two daughters of ViT, viz.\ gMLP~\cite{liu2021pay} (\figref{label:002}) and Swin Transformer~\cite{liu2021swin} (\figref{label:003}).
From \figref{label:005}, we may observe that gMLP’s overall structure is largely the same as ViT, but the structure of individual blocks has been significantly modified. On the other hand, Swin Transformer’s overall structure has been significantly modified in terms of the introduction of a hierarchical structure, but the internal structure of the individual blocks is largely the same as in ViT, except for the introduction of shifted windows.
Thus, there is little overlap and a high degree of independence in the differences from ViT in each of these methods. Therefore, by incorporating both at the same time, higher improvements can be expected.

Based on these observations, the authors propose \textit{gSwin} as a new vision model that inherits features of both gMLP and Swin Transformer. \figref{label:004} shows the architecture of gSwin. In the following subsection, we will discuss the methods of Window-SGU.
Note that this is not the same as MetaFormers~\cite{yu2021metaformer},
a generalization of Swin Transformer in which the self-attention module is replaced with other modules.
Although the use of gMLP module has already been considered in MetaFormers,
the architecture is not the same as ours,
in which
the subsequent LayerNorm, MLP and residual connections after each gMLP block are omitted.
Despite the fewer number of parameters, the performance of the present method was higher than that of MetaFormers,
which will be shown in \secref{label:013}.

\subsection{Spatial Gating Unit (SGU) and Window-SGU}\label{label:009}

Let us first recall the gMLP~\cite{liu2021pay} model. \figref{label:002} shows the overall architecture of gMLP,
which consists of stacking of axis-wise FCN (fully-connected nets),
LayerNorm,
GELU activation function,
and the spatial gating unit (SGU).
The structure of SGU is as follows.
Let $\mathbf{Z}$ be the input feature of size $H \times W \times 2C$,
and let $\mathbf{Z}_1, \mathbf{Z}_2$ $((HW) \times C)$ be the split features.
%$[\mathbf{Z}_1, \mathbf{Z}_2] \gets \mathrm{split}(\mathrm{reshape}(\mathbf{Z}))$.
Then the SGU performs the following operation,
\begin{equation}
    \text{SGU:}\quad
    \mathbf{Y} = \mathbf{Z}_1 \odot (\mathbf{W} \mathbf{Z}_2 + \bm{b}),
	\label{label:010}
\end{equation}
where $\mathbf{W} \in \mathbb{R}^{(HW) \times (HW)}$ and  $\bm{b} \in \mathbb{R}^{(HW)}$ are trainable parameters, and $\odot$ denotes the element-wise multiplication.
Such a gating mechanism allows us to explicitly incorporate
binomial relations between all the 2D site (pixel) pairs of the input feature,
and thus it could be an alternative mechanism to self-attention.

According to \cite[Fig.~3]{liu2021pay}, the resulting learned weight parameters $\mathbf{W}$ were almost-sparse Gabor-like patterns.
Furthermore, \cite[Fig.~9]{liu2021pay} suggests that Toeplitz-like (i.e., shift invariant) structures are automatically acquired.
These facts imply that it is not necessary to model $\mathbf{W}$ as a matrix with full degrees of freedom.
Instead, it would be sufficient to consider only local interactions by partitioning the features into small patches.

Based on this motivation,
this paper proposes the Window-SGU module;
unlike the SGU of gMLP, the Window-SGU considers the gating operation on small patches instead of the entire image.
That is, we first split the input feature $\textbf{Z}$ into small patches $\textbf{Z}^{(i)}$ of size $h \times w \times 2C$,
where $h<H$, $w<W$.
Then, after the patch partitioning, each patch is processed independently by the SGU module,
\begin{equation}
    \text{Window-SGU:} \quad
    \mathbf{Y}^{(i)} = \mathbf{Z}_1^{(i)} \odot (\mathbf{W}_\text{win} \mathbf{Z}_2^{(i)} + \bm{b}_\text{win}),
    \label{label:011}
\end{equation}
where the trainable parameters $\mathbf{W}_\text{win}$ and $\bm{b}_\text{win}$ are shared across all patches.
The resulting $\mathbf{Y}^{(i)}$-s are again tiled and the boundaries are processed appropriately
to obtain $\mathbf{Y}$ of size $H\times W \times C$.

\subsection{Shifted Window and Padding-free Shifted Window}
\begin{wrapfigure}[11]{r}{0.35\linewidth}
	\centering
	\vspace{-10pt}
	\includegraphics[width=0.95\linewidth]{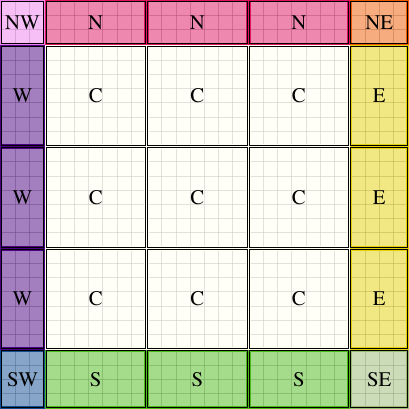}
	\vspace{-3mm}
	\caption{Padding-free Shifted Window}
	\label{label:012}
\end{wrapfigure}
As proposed in~\cite{liu2021swin},
the pattern of patch partitioning must be shifted from layer to layer,
in order to avoid block noise.
The shifted pattern naturally leads to the problem of unevenly shaped patches near the boundaries,
and some measures must be taken to handle them.
However, the simple zero-padding costs FLOPs,
and the cyclic shift~\cite{liu2021swin} cannot be used in the present case because SGU does not have masking capabilities.

Here, we propose another approach equivalent to zero-padding but more efficient;
the Padding-free shift.
Patches ``C'' of size $h \times w$
can fill the interior of the image,
but the boundaries need to be filled with sub-patches of smaller sizes e.g.\ ``NW'', ``N'', $\cdots$, as shown in~\figref{label:012}.
These sub-patches are grouped together with the ones of the same size,
and the group is processed by the SGU above.
The weight and bias of each group, e.g.\
$(\mathbf{W}^{\text{NW}}, \bm{b}^{\text{NW}}), (\mathbf{W}^{\text{N}}, \bm{b}^{\text{N}})$,
etc., share those of the main group ``C'', $(\mathbf{W}^{\text{C}}, \bm{b}^{\text{C}})$,
so that results are the same as if we had used zero-padding.
This Padding-free shift reduces FLOPs of gSwin-T from 3.8~G (zero-padding) to 3.6~G.

\section{Experiment}\label{label:013}

In this paper, following previous studies, we conducted evaluation experiments on the following three benchmark tasks:
image classification using ImageNet-1K~\cite{deng2009imagenet}, and object detection
using MS-COCO~\cite{lin2014microsoft}
and semantic segmentation using ADE20K~\cite{zhou2019semantic} as downstream tasks.
In our experiments, we compared three gSwin variants, details of which are shown in
the \#heads column in \secref{label:020}.
\tabref{label:014} shows the parameter setting for each variant.
We set window size to $7$ for all gSwin models.

\begin{table}[!t]
	\centering
	\caption{Parameter settings of our method.}%
    \label{label:014}%
    \label{label:015}%
	\vspace{-3mm}%
	{
	\begin{tabular}{@{}c@{\hskip 3mm}c@{\hskip 2mm}c@{\hskip 2mm}c@{\hskip 2mm}c@{\hskip 2mm}c@{\hskip 2mm}c@{}}
        \toprule
        \multirow{2}{*}{Method} & \multirow{2}{*}{$C$} & \multirow{2}{*}{\#layer} & \multirow{2}{*}{\#heads} &
		\multicolumn{3}{c}{Drop path rate} \\
		&&&&ImageNet & COCO & ADE20K\\
        \midrule
        gSwin-VT & $60$ & $\{2, 4, 10, 4\}$ & \ph$6$ & $0.25$   & $0.25$   & $0.2$\\
    	gSwin-T  & $64$ & $\{4, 4, 16, 4\}$ & $12$   & $0.35$   & $0.3$\ph & $0.3$\\
    	gSwin-S  & $72$ & $\{4, 4, 32, 4\}$ & $12$   & $0.5$\ph & $0.4$\ph & $0.4$\\
    	\bottomrule
    \end{tabular}
	}
\end{table}

\begin{table}[!t]
	\centering
	\caption{
		ImageNet-1K~\cite{deng2009imagenet} classification top-1 accuracy.
		(Digit in the parentheses indicates the standard error of the last digit.)
	}%
    \label{label:016}%
	\vspace{-3mm}%
	\begin{tabular}{@{}l@{}rrr@{}}
		\toprule
		Method          & \#param. ($\downarrow$) & FLOPs  ($\downarrow$) & Top-1 acc.  ($\uparrow$) \\
		\midrule
		DeiT-S~\cite{touvron2021training}
		& 22 M      & 4.6 G  & 79.8\PH\%     \\
		DeiT-B~\cite{touvron2021training}
		& 86 M      & 17.5 G & 81.8\PH\%     \\
		Swin-T~\cite{liu2021swin}
		& 28 M      & 4.5 G  & 81.29(3)\%     \\
		Swin-S~\cite{liu2021swin}
		& 50 M      & 8.7 G  & 83.02(7)\%     \\
		\midrule
		gMLP-Ti~\cite{liu2021pay}
		& 6 M       & 1.4 G  & 72.3\PH\%     \\
		gMLP-S\cite{liu2021pay}
		& 20 M      & 4.5 G  & 79.6\PH\%     \\
		gMLP-B\cite{liu2021pay}
		& 73 M      & 15.8 G & 81.6\PH\%     \\
		Hire-MLP-Ti\cite{guo2021hire}
		& 18 M      & 2.1 G  & 79.7\PH\%     \\
		Hire-MLP-S~\cite{guo2021hire}
		& 33 M      & 4.2 G  & 82.1\PH\%     \\
		Hire-MLP-B\cite{guo2021hire}
		& 58 M      & 8.1 G  & 83.2\PH\%     \\
		\midrule
		PoolFormer-S12~\cite{yu2021metaformer}
		& 12 M      & 2.0 G  & 77.2\PH\%     \\
		PoolFormer-S24~\cite{yu2021metaformer}
		& 21 M      & 3.6 G  & 80.3\PH\%     \\
		PoolFormer-S36~\cite{yu2021metaformer}
		& 31 M      & 5.2 G  & 81.4\PH\%     \\
		PoolFormer-M36~\cite{yu2021metaformer}
		& 56 M      & 9.1 G  & 82.1\PH\%     \\
		PoolFormer-M48~\cite{yu2021metaformer}
		& 73 M      & 11.9 G & 82.5\PH\%     \\
		\midrule
		gSwin-VT (ours) & 16 M      & 2.3 G  & 80.32(1)\%     \\
		gSwin-T (ours)  & 22 M      & 3.6 G  & 81.71(5)\%     \\
		gSwin-S (ours)  & 40 M      & 7.0 G  & 83.01(4)\%     \\
		\bottomrule
	\end{tabular}
\end{table}
\begin{table}[!t]
	\centering
	\caption{
		COCO~\cite{lin2014microsoft} object detection and instance segmentation.
	}
	\label{label:017}
	\vspace{-3mm}%
	\begin{tabular}{@{}l@{}rrr@{}}
		\toprule
		Method          & \#param. ($\downarrow$)
		& box-AP  ($\uparrow$) &
		%$\text{AP}^{\text{mask}}$  ($\uparrow$) \\ \midrule
		mask-AP ($\uparrow$) \\ \midrule
		Swin-T~\cite{liu2021swin}
		& 86 M      & 50.52(5)\% & 43.78(6)\% \\
		Swin-S~\cite{liu2021swin}
		& 107 M     & 51.98(3)\% & 44.99(6)\% \\ \midrule
		PoolFormer-S12~\cite{yu2021metaformer}
		& 32 M & 37.3\PH\% & 34.6\PH\% \\
		PoolFormer-S24~\cite{yu2021metaformer}
		& 41 M & 40.1\PH\% & 37.0\PH\% \\
		PoolFormer-S36~\cite{yu2021metaformer}
		& 51 M & 41.0\PH\% & 37.7\PH\% \\ \midrule
		gSwin-VT (ours) & 73 M      & 49.48(2)\% & 42.87(2)\%   \\
		gSwin-T (ours)  & 79 M      & 50.97(2)\% & 44.16(3)\%   \\
		gSwin-S (ours)  & 97 M      & 52.00(7)\% & 45.03(5)\%   \\
		\bottomrule
	\end{tabular}
\end{table}

\subsection{Image Classification (ImageNet $224 \times 224$)}

The training conditions were largaly aligned with those of Swin Transformer~\cite{liu2021swin}.
The optimizer was AdamW
($\text{lr} = 10^{-3}$, weight decay $= 0.05$.)
Data augmentation and regularization techniques
were adopted including
randaugment~\cite{cubuk2020randaugment},
and the drop-path regularization~\cite{larsson2016fractalnet},
whose rates are shown in \tabref{label:015} (tuned by grid search).
The batch size was 1024 and the number of epochs was 300 with cosine scheduler (first 20 epochs for linear warm-up).
Of these, the checkpoint with the best top~1 accuracy in the validation dataset was selected.
The training was performed twice with different seeds.
We report the average (and unbiased variance) values of the two trials.

We compared the image recognition performance of our method with that of
DeiT~\cite{touvron2021training}, Swin Transformer~\cite{liu2021swin}, gMLP~\cite{liu2021pay}
and Hire-MLP~\cite{guo2021hire}.
\tabref{label:016} shows the results.
Compared to Swin Transformers, gSwin-T achieves $+0.4$\% top-1 accuracy with 21\% less parameters,
and gSwin-S achieves the same accuracy with 20\% less.
We may observe that gSwin (proposed) is more efficient in terms of the number of parameters and floating point
operations (FLOPs) to achieve the same level of accuracy with others.

\subsection{Object Detection (COCO)}
We next performed the object detection and the instance segmentation experiments.
The evaluation metric was the box-AP (average precision) and the mask-AP.
We used the Cascade Mask R-CNN~\cite{He2017MaskR, Cai2018CascadeRD} as object detection framework
following~\cite{liu2021swin}.

Rather than training a new model in a full-scratch fashion,
we reused the two best checkpoints above.
Starting from each checkpoint,
we continued training for 36 epochs where the batch size was 16.
The optimizer was AdamW
($\text{lr}=10^{-4}, \text{decay} = 0.05$).
During training, the multi-scale learning was performed,
where input images were resized to have short sides of 480--800 pixels and long sides of up to 1333 pixels.
For every epoch, we evaluated the AP scores, and after the training, we selected the best checkpoint with the highest box-AP.
This transfer learning was done three times using different random seeds,
which resulted in 6 different models. We report the average score of these 6 models.

From results in \tabref{label:017}, we observe that gSwin-T achieves
$+0.45/+0.38$ box/mask-AP with 8\% less parameters, and gSwin-S achieves similar APs with 9\% less parameters.

\subsection{Semantic Segmentation (ADE20K $512 \times 512$)}
We also performed an experiment on the image segmentation task.
The evaluation metric was mIoU (mean intersection over union).
We used UPerNet~\cite{xiao2018unified} as base framework, as in \cite{liu2021swin}.

This experiment is also based on the
transfer learning from the best checkpoints of ImageNet-1K.
Starting from the best checkpoints,
we continued training for 160k steps with linear scheduler (first 1.5k steps for linear warm-up)
where the batch size was 16.
The optimizer was AdamW
($\text{lr}=6\times 10^{-5}, \text{decay} = 0.01$).
The augmentations we adopted was random horizontal flip, random rescaling
(the resolution ratio within the range from $\times 0.5$ to $\times 2.0$) and random photometric distortion, as in \cite{liu2021swin}.
During the training, we evaluated the mIoU score for every 4k steps,
and after the training, we selected the best checkpoint with the highest validation score.
We similarly performed this transfer learning three times using different random seeds for two different checkpoints,
which resulted in 6 different models. We report the average score of these 6 models.

Results on ADE20-K are shown in \tabref{label:018}, where the mIoU(aug)
column is the score with test-time multi-scale (with factors 0.5, 0.75, 1.0, 1.25, 1.5, 1.75) horizontal-flip augmentation.
These show that gSwin-T achieves $+1.8/+1.85$\% mIoU with/without augmentation with 13\% less parameters,
and gSwin-S achieves $+0.46/+0.57$\% mIoU with 14\% less parameters, compared to Swin Transformer.

\begin{table}[!t]
	\caption{
		ADE20K~\cite{zhou2019semantic} semantic segmentation.
	}
	\label{label:018}
	\vspace{-3mm}%
	\begin{tabular}{@{}l@{}rrr@{}}
		\toprule
		method          & \#param. ($\downarrow$) & mIoU ($\uparrow$)   & mIoU(aug) ($\uparrow$) \\ \midrule
		Swin-T~\cite{liu2021swin}
		& 60 M      & 44.3(1)\% & 45.8(1)\% \\
		Swin-S~\cite{liu2021swin}
		& 81 M      & 47.7(1)\% & 49.1(1)\% \\ \midrule
		PoolFormer-S12~\cite{yu2021metaformer}
		& 16 M & 37.2\Ph\% & \\
		PoolFormer-S24~\cite{yu2021metaformer}
		& 23 M & 40.3\Ph\% & \\
		PoolFormer-S36~\cite{yu2021metaformer}
		& 35 M & 42.0\Ph\% & \\
		PoolFormer-M36~\cite{yu2021metaformer}
		& 60 M & 42.4\Ph\% & \\
		PoolFormer-M48~\cite{yu2021metaformer}
		& 77 M & 42.7\Ph\% & \\ \midrule
		gSwin-VT (ours) & 45 M      & 43.4(1)\% & 45.1(1)\%   \\
		gSwin-T (ours)  & 52 M      & 46.1(1)\% & 47.6(1)\%   \\
		gSwin-S (ours)  & 70 M      & 48.2(1)\% & 49.7(1)\%   \\
		\bottomrule
	\end{tabular}
\end{table}

\section{Ablation Studies}
\begin{table}[!t]
	\centering
	{
	\caption{
		Ablation study on $K$ (number of heads) and RPB (relative position bias) in gSwin-T.
		The asterisk * indicates that only one checkpoint for ImageNet-1K was trained.
	}
	\label{label:019}
	\vspace{-3mm}%
	\begin{tabular}{@{}ccc|r|rr@{}}
		\toprule
		 &        &              & ImageNet                   & \multicolumn{2}{c}{ADE20K} \\
		 & $K$    & RPB          & Top-1 ($\uparrow$)         & mIoU      & mIoU(aug) \\ \midrule
		*& \ph{}1 & \cmark & 80.91\phantom{(00)}\%      & 43.8(1)\% & 45.3(1)\% \\
		*& \ph{}3 & \cmark & 81.46\phantom{(00)}\%      & 45.5(3)\% & 47.2(3)\% \\
		 & \ph{}6 & \cmark & 81.61(1)\phantom{0}\%      & 45.8(1)\% & 47.2(1)\% \\
		 & 12     & \cmark & 81.71(5)\phantom{0}\%      & 46.1(1)\% & 47.6(1)\% \\
		*& 12     & \xmark & 81.27\phantom{(00)}\%   & 44.8(1)\% & 46.5(1)\% \\
		 & 24     & \cmark & 81.80(9)\phantom{0}\%      & 46.2(1)\% & 47.7(1)\% \\
		 & 48     & \cmark & 81.78(10)\%  & 45.7(1)\%   & 47.1(1)\% \\
		\bottomrule
	\end{tabular}
	}
\end{table}

\subsection{Multi-Head SGU}\label{label:020}

Although a single weight matrix $\mathbf{W} \in \mathbb{R}^{(WH)\times(WH)}$ was used in \myeqref{label:010}~\cite{liu2021pay},
we consider to adopt the \textit{multi-head} structure
which improves the accuracy of Transformer-based models~\cite{vaswani2017attention}.
In the multi-head SGU, instead of the matrix, the weight and bias tensors are introduced,
i.e.\ the weight $\mathbf{W}_\text{win}$ of size $(wh) \times (wh) \times K$
and the bias $\mathbf{b}_\text{win}$ of size $(wh) \times K$,
where $K$ being the number of heads, which should be a divisor of $C$.
The Window-SGU~\myeqref{label:011} is modified as follows,
\begin{gather}
    \mathbf{Y}^{(i)}[:,:,c]
    =
	\mathbf{Z}_1^{(i)}[:,:,c] \odot \left(
		\mathbf{W}_\text{win}[:,:,k] \mathbf{Z}_2^{(i)}[:,:,c]
		+\mathbf{b}_\text{win}[:,k]
		\right), \nonumber \\
	\qquad\text{where}\quad c \in \{1, 2, \cdots, C\}, \quad k = \lceil cK/C \rceil.
\end{gather}
The parameters to be learned increase linearly as $K$ increases,
but the FLOPs are not affected by $K$.
Increasing $K$ allows the model to interact tokens in more complicated patterns
at once and helps to keep the model capacity to represent complex spatial interactions.

We did ablation studies on the choice of $K$ using gSwin-T model.
\tabref{label:019} shows the results for $K=1,3,6,12,24,48$.
Firstly, we found that the single-head model ($K=1$) is less effective than multi-head models.
The accuracy increases as $K$ increases, and it saturates at $K\sim 12$.
From these results, we chose $K=12$ for gSwin-T and gSwin-S
($K=6$ for gSwin-VT, to keep model small).

\subsection{Relative position bias}
\begin{figure}
		\centering
		\def\mywidth{0.45\linewidth}
		\subfloat[With RPB.]{
			\includegraphics[width=\mywidth]{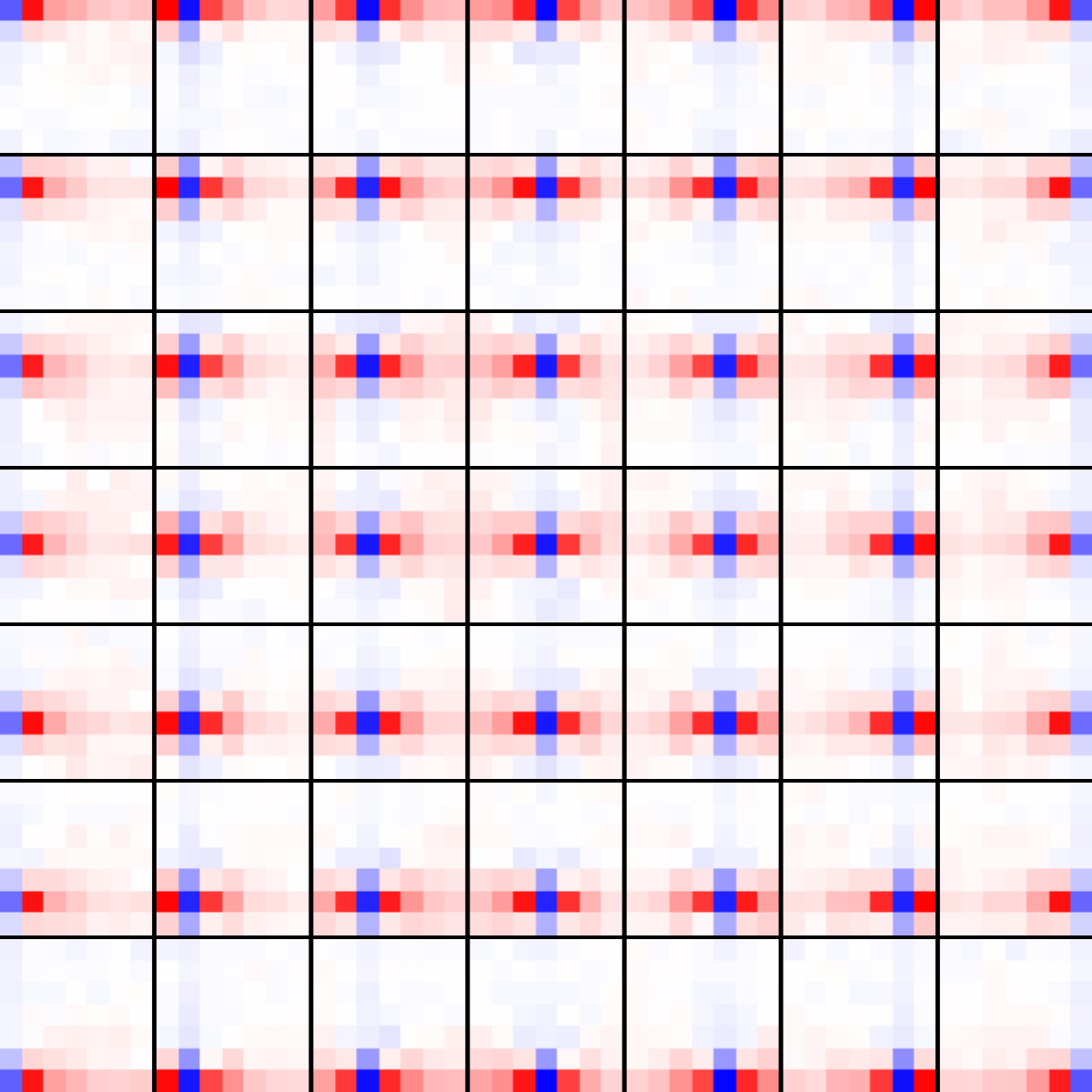}
			\label{label:021}
		}
		\subfloat[Without RPB.]{
			\includegraphics[width=\mywidth]{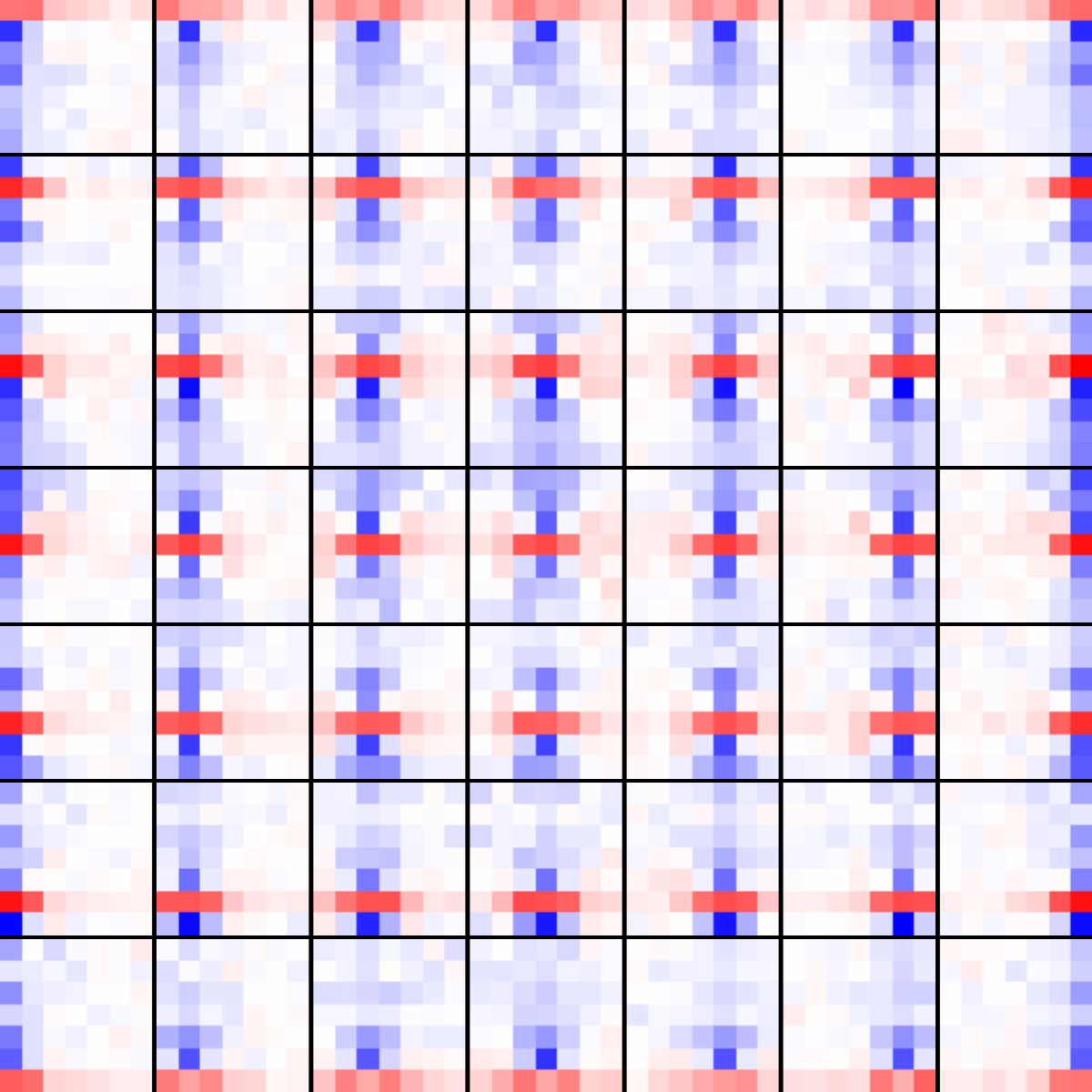}
			\label{label:022}
		}
		\caption{A head of the 11th layer of the 3rd block of gSwin-T with and without relative positional bias (RPB).
		In both cases, nearly shift-invariant weights were acquired by the training.}
		\label{label:023}
	\end{figure}

The relative positional bias (RPB) is a technique
that aims to improve performance and efficiency by adding a bias $\mathbf{B}$, which has information about the spatial relations between tokens,
to the attention score as a kind of prior knowledge when calculating the attention, e.g.\
$\textbf{A} = \text{softmax}(\mathbf{Q}\mathbf{K}^{\mathsf{T}}/\sqrt{d} + \mathbf{B})\textbf{V}$,
and it has been reported to improve performance when applied to the Swin Transformer~\cite{liu2021swin}.
Such a prior is essentially necessary to explicitly inform the Transformer-based architectures about the spatial structure of data.

On the other hand, as discussed in \secref{label:009},
gMLP~\cite{liu2021pay} can learn
such spatial interactions, and such a guidance is not necessary.
We have experimentally confirmed it is also the case for gSwin.
We introduced the RPB mechanism to gSwin by modifying $\mathbf{W}$ in the SGU to
$\mathbf{W}=\mathbf{W}'+\mathbf{B}$,
where each element of the bias matrix $\mathbf{B}$ is only dependent on relative positions between tokens.
We observed that gSwin can be trained without this term.
\figref{label:023} shows excerpted weights $\mathbf{W}$
with or without the RPB term, and we may observe that nearly shift-invariant weights are acquired in both cases.

Nevertheless, in practice, RPB helped gSwin to achieve better accuracy
as shown in $K=12$ rows in \tabref{label:019},
with negligible increase in model size.

\section{Conclusion}\label{label:024}
In this paper, we proposed the gSwin,
which fuses two visual models: one that incorporates the spatial hierarchical structure of the input feature map,
and the other that uses a parameter-efficient MLP architecture as an alternative for the self-attention mechanism.
Specifically, it incorporates the hierarchical Swin (shifted-window) structure to the multi-head gMLP.
Despite the very simple idea of integrating two existing ideas, promising experimental results were obtained;
the present method outperformed Swin Transformer on
three common benchmark tasks of vision, i.e.\ image classification (ImageNet-1K),
object detection (COCO), and semantic segmentation (ADE20K).
Therefore, the gSwin could also be a promising candidate as a basis for other downstream tasks just as Swin Transformer is.
Future studies will include the examination of the effectiveness of the gSwin-based methods for specific downstream tasks.
In addition, further studies would also be needed on a better structure of the Vision MLPs.

\section{Acknowledgments}
This paper is based on results obtained from a project subsidized by the New Energy and Industrial Technology Development Organization (NEDO).

{
\bibliographystyle{IEEEbib}
\bibliography{main_modified}
}
\end{document}